\documentclass[letterpaper, 10 pt, conference]{ieeeconf}  

\IEEEoverridecommandlockouts                              

\usepackage{graphics} 
\usepackage{epsfig} 
\usepackage{mathptmx} 
\usepackage{times} 
\usepackage{amsmath} 
\usepackage{amssymb}  
\usepackage{cite} 

\title{\LARGE \bf
Planning for the Unexpected: Explicitly Optimizing Motions for Ground Uncertainty in Running
}

\author{Kevin Green, Ross L. Hatton, Jonathan Hurst
\thanks{This work was supported by DARPA contract W911NF-16-1-0002, NSF Grant No. 1462555-CMM and NSF Grant No. 1314109-DGE. \newline
\- All authors are with the School of Mechanical, Industrial, \& Manufacturing Engineering at Oregon State University, Corvallis, OR, USA. Email
        {\tt\small \{greenkev, ross.hatton, jonathan.hurst\}@oregonstate.edu }}%
}

\begin{document}

\maketitle
\thispagestyle{empty}
\pagestyle{empty}

\begin{abstract}

We  propose a method to generate actuation plans for a reduced order, dynamic model of bipedal running. 
This method explicitly enforces robustness to ground uncertainty. 
The plan generated is not a fixed body trajectory that is aggressively stabilized: instead, the plan interacts with the passive dynamics of the reduced order model to create emergent robustness. 
The goal is to create plans for legged robots that will be robust to imperfect perception of the environment, and to work with dynamics that are too complex to optimize in real-time. 
Working within this dynamic model of legged locomotion, we optimize a set of disturbance cases together with the nominal case, all with linked inputs. 
The input linking is nontrivial due to the hybrid dynamics of the running model but our solution is effective and has analytical gradients.
The optimization procedure proposed is significantly slower than a standard trajectory optimization, but results in robust gaits that reject disturbances extremely effectively without any replanning required.

\end{abstract}

\section{Introduction}

Dynamic locomotion such as running and walking has many dimensions beyond position trajectories, which are merely one symptom of the resulting behavior. 
As such, new approaches are needed to incorporate powerful existing motion planning and control methods with the dynamic behaviors of legged locomotion. 
Complicating factors include underactuation, nonlinear hybrid dynamics, large system dimensionality and significant uncertainties in ground properties. 
However, legged locomotion is not so complex as it first appears, because most behaviors can be described by relatively simple reduced-order models, showing some promise for planning within this dynamic space. 
Many reduced-order models consist of a point mass body and a massless leg that can apply forces from a contact point toward the point mass, where body motion is only influenced by gravity and the forces applied by the leg. 
Examples of this type of model include the inverted pendulum (IP) model, the linear inverted pendulum (LIP) model, the spring loaded inverted pendulum (SLIP) model, and the actuated spring loaded inverted pendulum (ASLIP) model. 
The differentiating factor between these models is the calculation of the applied leg force.

\begin{figure}
  \centering
  \includegraphics[width = 0.98\columnwidth]{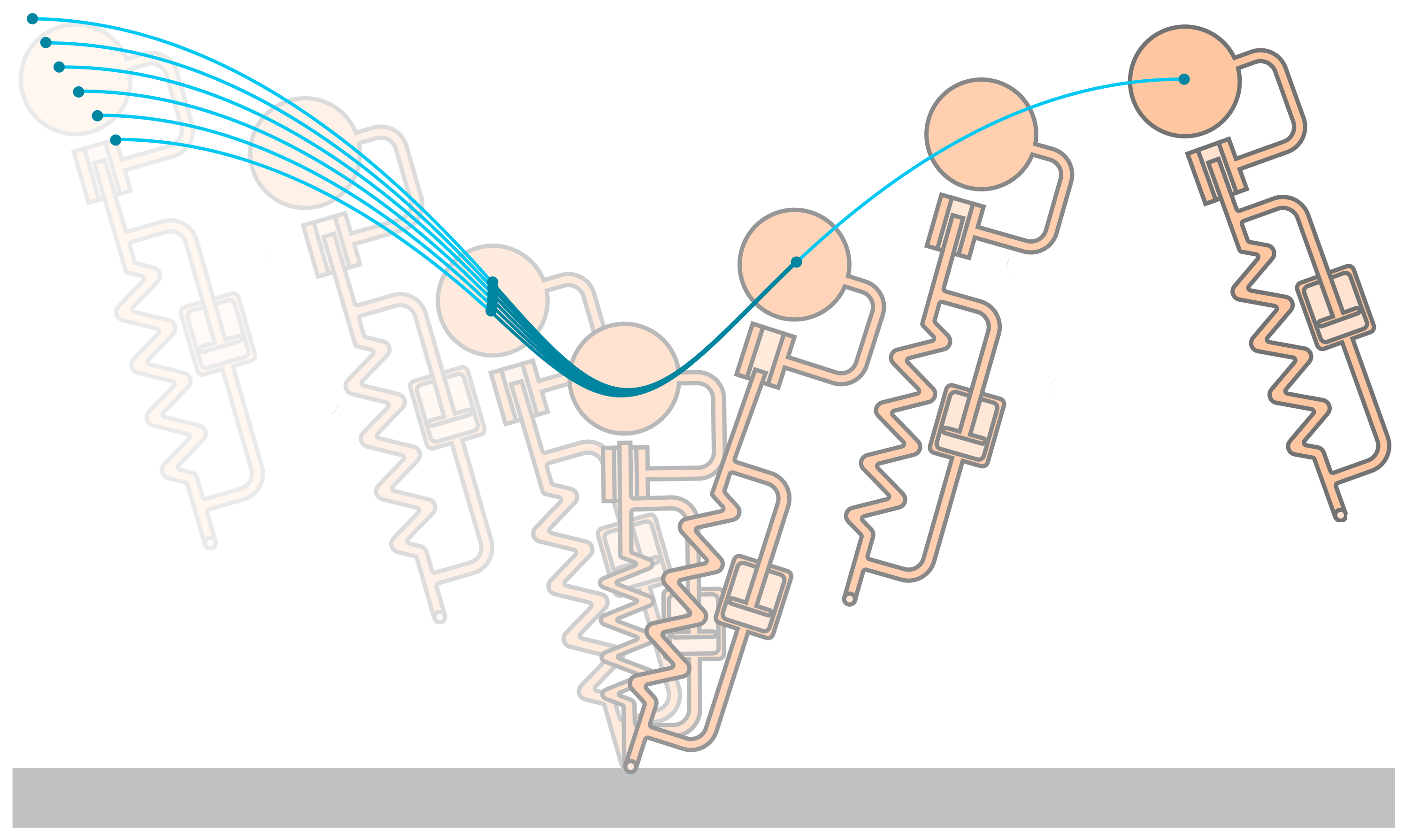} 
  \caption{A robust motion plan for the actuated SLIP model. All trajectories use the same control inputs, yet they all converge from different heights to the same final apex state.}
\label{fig:aaslipCycle}
\end{figure}

When walking and running, the ground height for the next step cannot be measured perfectly so intrinsic robustness to errors in ground height is extremely desirable. 
Ground sensing is a difficult problem because of the complex dynamics of legged locomotion\cite{Cadena16tro-SLAMfuture}. Feet can slip which complicates proprioceptive estimation \cite{Bloesch2012}. 
Cameras, LIDAR and other sensors experience difficult-to-predict motion throughout the gait cycle. 
Due to the sharp cost of failures (falls and subsequent damage) it is desirable to be robust to the uncertainties in ground height. 
This is the perspective that motivates much of the work on generating robust blind locomotion \cite{Ramezani2013}. 

In this work we utilize the ASLIP model as a template for planning robust running motion because of its demonstrated robustness and its rich actuation space. 
This model has a leg that consists of an extensible actuator in series with a lightly damped spring. With a simple open loop actuator trajectory this model rejects ground height disturbances when hopping vertically \cite{Rizzi2007}. 
This leads to the question, "How robust could this model be with the best choice of open loop actuator trajectory?"
We developed a method of explicitly solving for motions that reject ground height disturbances without sensing, replanning and reacting.
Our approach is to optimize the expected motion for a set of disturbance cases in one large problem. 
This is not trivial because it requires linking the inputs between the different cases including accounting for the timing of the hybrid transitions. 
Our solution to input linking is effective and includes analytical gradients throughout. We provide an overview of relevant existing work in Section \ref{sec:background}. 
The ASLIP model and its hybrid dynamics are defined in Section \ref{sec:model}. 
In Section \ref{sec:methods} we describe the two trajectory optimization techniques we are comparing and the simulation we use to test performance. 
The results of both optimizations and the simulation testing are reported in Section \ref{sec:results}. 
Closing remarks are in Section \ref{sec:conclusions}.

\section{Background}
\label{sec:background}

Studying reduced order models has provided insight into the phenomenon of legged locomotion and how to create dynamic walking and running in robots.
The passive SLIP model explains most of the effects observed in human ground reaction forces during running and walking \cite{Geyer2006}.
If this model is extended with swing leg dynamics, it can generate all common bipedal gaits \cite{Gan2018BipedalGaits}.
Further the SLIP model can be used to generate foot placement policies that regulate forward speed \cite{Ernst2009}.
In the actuated and damped ASLIP model, an open loop cyclic reference trajectory was shown to reject both ground height and ground impedance variations when hopping vertically \cite{Rizzi2007}.
Preflex (pre-reflex) behaviors on an extension of the ASLIP model can aid in mitigating sensing delays when the system encounters disturbances \cite{Hubicki2015b}.
These discoveries inform us on how to create and stabilize legged locomotion.

The insights from reduced order models have been successfully leveraged in generating control methods for legged robots.
The ATRIAS robot successfully demonstrated robust blind walking by leveraging knowledge from reduced order models about foot placement, energy injection and clock based open-loop feed-forward signals \cite{Hubicki2018ATRIAS}.
The RHex hexapod robot was stabilized with a feedback policy from a clock-driven SLIP \cite{altendorfer2004stability}.

Other control approaches directly manipulate reduced order models to plan motions online in a model predictive control approach.
A common approach for bipedal locomotion is it use the linear inverted pendulum model due to its linear dynamics \cite{Feng2014, dimitrov2011sparse, Apgar2018}.
This allows the robot to quickly reason about where to place its feet given the estimated state and the goals from a high level planner or an operator.




\section{Dynamic Model}
\label{sec:model}

\begin{figure}[t]
  \centering
  \includegraphics[width = 0.7\columnwidth]{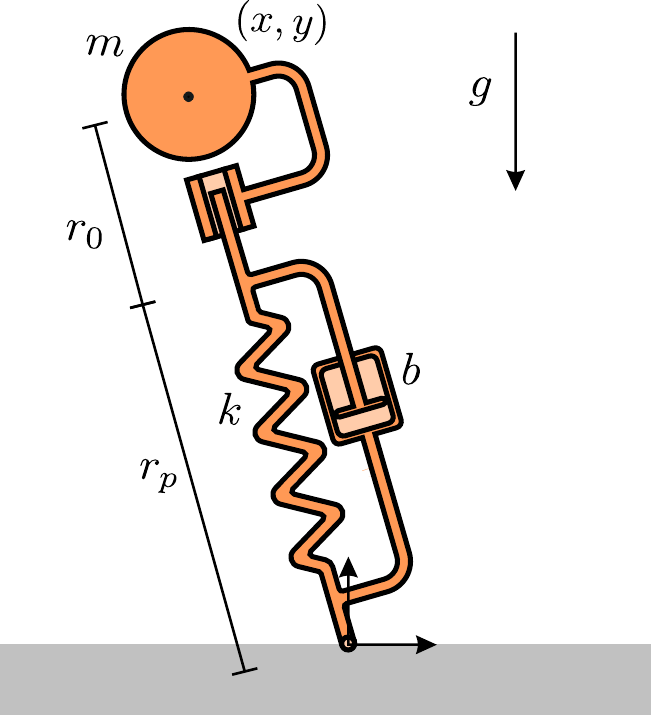}
  \caption{The ASLIP model in stance with labeled parameters and state variables. The origin for the body position $(x,y)$ is the contact point.}
\label{fig:LabeledASLIP}
\end{figure}

The ASLIP model consist of a point mass body with mass $m$ and no rotational inertia as well as a massless leg.
This leg has a linear extension actuator that is assumed to be a rigid position input.
The output of this actuator is connected to the massless point foot through a damped linear spring with stiffness $k$ and viscous damping $b$. 
This system has states corresponding to the body position $x$, $y$ and velocity $\dot{x}$, $\dot{y}$, the leg actuator set point position $r_0$ and velocity $\dot{r}_0$, and the passive spring deformation $r_p$. 
The leg actuation extension is limited to be between $l_0$ and $l_0/2$ where $l_0$ is used as a descriptive length of the leg.
Gravitational acceleration is $g$.
These coordinates and parameters are labeled on the system diagram in Fig. \ref{fig:LabeledASLIP}.
We include the actuator velocity as a state because we use the acceleration of the set point $\ddot{r}_0$ as the control input. 
The commanded acceleration is limited in magnitude as a proxy for absolute torque limits.
Here $5g$ is used as the maximum acceleration
The passive spring deflection must be a part of the state because during flight phase the spring and damper create first order dynamics.
This model has both a discrete control action in the foot placement as well as a continuous control action in the leg extension actuator which results in a much richer action space than the passive SLIP model.

\subsection{Equations of Motion}
This system has two distinct dynamic modes, flight and stance. 
In flight phase the dynamics of the body and the leg are fully decoupled. 
The body exhibits ballistic motion with the equations of motion
\begin{align}
\ddot{x} &= 0 \\
\ddot{y} &= -g.
\end{align}
The leg set point motion is only influenced by the commanded acceleration ($\ddot{r}_{cmd}$) because it is assumed to be rigidly actuated,
\begin{equation}
\ddot{r}_0 = \ddot{r}_{cmd}
\label{eq:setPointIntegration}
\end{equation}
The passive deflection of the spring has a first order response during flight due to the lack of foot mass
\begin{equation}
\dot{r}_p = -\frac{k}{b} r_p
\end{equation}

In stance phase the toe is constrained to stay in contact with the ground and leg is able to apply forces on the main body, but only in the leg length direction.
In the application of this constraint, the spring deflection variable ($r_p$) is made dependent on the body and set point position.
To simplify the description of the dynamics, the origin of the body’s position coordinate system is set to be the foot contact point.
This makes the spring deflection and velocity
\begin{align}
r_p &= r - r_0\\
\dot{r}_p &= \dot{r} - \dot{r}_0,
\end{align}
where $r$ is the total leg length and $\dot{r}$ is the total leg velocity. 
The stance center of mass dynamics are
\begin{align}
\ddot{x} &= \frac{x F}{m r} \\
\ddot{y} &= \frac{y F}{m r} -g
\end{align}
where $F$ is the leg force on the body, defined as
\begin{equation}
F = k(r_0 - r) + b(\dot{r}_0 - \dot{r})
.\end{equation}
We assumed that the set point is a rigid position source so it is influenced only by the $\ddot{r}_{cmd}$ control signal despite the external loads it is supporting in stance.

\subsection{Hybrid Transition Model}
\label{sub:transition}
We are using this model to represent the sagittal plane dynamics of bipedal aerial running. 
This means that the robot will cycle through the phases of flight, left leg stance, flight and right leg stance before repeating. 
Given symmetries in the sagittal plane we can analyze only half of this cycle. 
The start and end points of this half cycle are somewhat arbitrary, but a common choice is to use the apex condition in flight as the start and end point \cite{Geyer2006}. 
With this start and end point, the phases of our half cycle are descending flight, stance, ascending flight.

Touchdown is when the model transitions from flight to stance which occurs when the foot reaches the ground.
This is more precisely described as the states at the moment of touchdown intersect the guard surface
\begin{equation}
\sqrt{ x^2 + y^2} = r_0 + r_p.
\end{equation}

Liftoff is where the model transitions from stance phase to flight phase when the ground reaction force goes to zero. 
This is not the same as when the spring has zero deflection because of the damping in the spring. 
The liftoff transition guard is described as having zero force in the spring
\begin{equation}
k(r_0 - r) + b(\dot{r}_0 - \dot{r}) = 0.
\end{equation}
This model does not include the case where the force vector exits the friction cone and causes the foot to slip.

\subsection{Nondimensionalization}

If this model was being used to generate motions for a specific robot, one would use meaningful physical parameters from that robot for the reduced order model.
Here we are interested in results that are generally applicable so we nondimensionalize our model.
The nondimensionalization of our ASLIP model is based on previous work nondimensionalizing SLIP models and variations on SLIP models \cite{hof1996scaling, Cnops2015}.
The characteristic units are the maximum leg set point length, mass of body and gravitational acceleration.
All parameters and states are represented relative to these quantities and are summarized in Table \ref{tab:parameters}.
Two parameters must be chosen, the leg stiffness and the leg damping.
The numbers we selected are similar to previous SLIP modeling papers \cite{Cnops2015} and are based on observations of human biomechanics \cite{alexander1988elastic}.
The leg stiffness we used is $20$ [$m g / l_0$] and the leg damping is $0.89$ [$m \sqrt{g/l_0}$].
The damping value is such that the body and leg system has a damping ratio of $0.1$ in stance. 

\begin{table}
  \begin{tabular}{ l  c  l    l }
                    & Symbol & Value & Description \\ \hline
    Base Units      & $m$         & $1$        [$m$]                 & Mass\\
                    & $l_0$       & $1$        [$l_0$]               & Max set point Length\\
                    & $g$         & $1$        [$g$]                 & Gravitational acceleration\\
    \hline
    Parameters      & $k$         & $20$       [$m g / l_0$]         & Leg spring stiffness\\
                    & $b$         & $0.89$     [$m \sqrt{g/l_0}$]    & Leg spring damping\\
    \hline
    States          &$x$         & -        [$l_0$]                 & Horizontal position\\
                    & $y$         & -        [$l_0$]                 & Vertical position\\
                    & $\dot{x}$   & -        [$\sqrt{g l_0}$]                 & Horizontal velocity\\
                    & $\dot{y}$   & -        [$\sqrt{g l_0}$]                 & Vertical velocity\\
                    & $r_0$         & -        [$l_0$]                 & Leg set point length\\
                    & $\dot{r}_0$   & -        [$\sqrt{g l_0}$]                 & Leg set point velocity\\
                    & $r_p$       & -        [$l_0$]                 & Leg spring deflection\\
    \hline
    Inputs          &$\ddot{r}_{cmd}$       & -        [$g$]                 & Leg set point acceleration\\
    
  \end{tabular}
  \caption {Nondimensional system parameters, states and inputs.}\label{tab:parameters}
\end{table}

\section{Methods} 
\label{sec:methods}
As a standard of comparison we first present a conventional minimal effort trajectory optimization.
Then we describe the explicitly robust trajectory optimization method we propose.
Finally, we describe a separate testing simulation to analyze the disturbance rejection capabilities of the open loop motion plans produced by these two optimization methods.

\subsection{Minimum Effort Optimization}

The minimum effort optimization seeks to find a single state trajectory and input signal that will result in moving from an initial apex state to the following apex state while minimizing a measure of actuator effort.
This is formulated as a direct collocation problem.
In this technique, the optimizer has access to both the discretized states and control inputs as decision variables as well as the time between the discritizations.
It is conventional to have the states evenly spaced in time with a single duration decision variable.
The dynamics are imposed as constraints between subsequent states and their inputs based on numeric integration techniques.
In this work we use trapezoidal integration.
Much more detail on this approach can be found in \cite{kellyTrajOpt}.

Our optimization is complicated by the three separate dynamic phases.
Each phase of the dynamics is implemented with its own set of discritized state and input variables and phase duration variable.
The final state of one phases is constrained to match the first state of the next phases.
Additionally the final states of the first flight phase and the stance phase must be at the hybrid transition guards described above in Section \ref{sub:transition}.

The initial height and forward velocity as well as the final height and forward velocity are constrained to match user-specified values.
To ensure the initial and final states are apex states, the vertical velocities are constrained to be zero.

The objective ($J$) we use is the integral of the set point acceleration squared,
\begin{equation}
    J = \int_0^\tau \ddot{r}_{\hspace{1 pt}0}^{\hspace{1 pt}2} \hspace{2 pt} dt 
\end{equation}
where $\tau$ is the total duration of the motion.
This is a useful objective both theoretically and practically.
If this was a real system where the set point actuator is a geared electric motor, this objective is proportional to the thermal energetic losses in the motor due to accelerating the actuator inertia \cite{YevOptimalControl}.
Practically this smooths the acceleration commands which increases the accuracy of the trapezoidal integration scheme \cite{kellyTrajOpt}. 

The constraints and objective functions are generated using a modification of COALESCE, a MATLAB based optimization problem generation library \cite{Jones2014}.
We generated the constraints, constraint Jacobian, objective and objective gradients analytically while preserving their sparsity.
This produces a nonlinear programming (NLP) problem that can be solved to local optimality using an off the shelf nonlinear solver.
We used IPOPT (an interior point technique package) to solve the NLP, but other implementations or optimization techniques could be used \cite{Wacher2006}.

\subsection{Disturbance Aware Trajectory Optimization}

The method we use to optimize for variations in ground height is an extension of the minimum effort technique presented above.
We not only create all the state variables and inputs for the expected motion, but also for some number of disturbance cases with different initial conditions.
All of the disturbance cases are thought of as different versions of what may happen during the planned step; this means that they must all share the same set point motion.
This is ensured through an input linking opteration described in the following section.
Each disturbance case still has the same final state constraints, forcing the optimizer to try to funnel each of the disturbance cases to the single final state.

The minimum effort objective was removed for this trajectory optimization because it was found to prevent the convergence of the optimization.
This makes this optimization problem more accurately a constraint satisfaction problem.
The limits on the maximum acceleration of the set point ensure that even without an explicit objective that the motion of the set point is relatively smooth.

One subtle aspect of this problem is that the disturbance cases are able to each select a different leg touchdown angle.
This may appear to conflict with the assumption that the model cannot know which disturbance case is occurring, but the optimizer is implicitly selecting a leg swing trajectory.
Each disturbance case contacts the ground at a different time at a different leg angle.
This set of leg angles over time exactly constitutes a leg swing retraction policy.



\subsubsection{Multiple Phase Input Linking}
The difficult aspect of working with all the disturbance cases together is that their control inputs must be linked together.
The system will not know which of the disturbance cases it is in so the inputs as a function of total time must match. 
Each disturbance case has flight and stance phases that take different lengths of time, so we cannot rely on the indexing of the collocation nodes to link instances. 
The solution is to use an additional set of decision variables evenly space through time that are not linked to any specific dynamic phase or collocation node.
Each of the collocation node inputs are constrained to be equal to the linear interpolated value from these control points.

To describe this constraint and its gradient, we first define a generic linear interpolation function and a zero order hold function.
The linear interpolation function (with extrapolation) as it is conventionally understood is
\begin{equation}
\text{LI}(x,v,x_q) = 
\begin{cases} 
      \frac{x_{2} - x_q}{x_{2} - x_1}v_1 + \frac{x_q - x_1}{x_{2} - x_1}v_{2} & x_q \leq x_2 \\
      \frac{x_{i+1} - x_q}{x_{i+1} - x_i}v_i + \frac{x_q - x_i}{x_{i} - x_i}v_{i+1} & x_{i} < x_q \leq x_{i+1}, \\
      & 2 < i < N-2 \\
      \frac{x_{N} - x_q}{x_{N} - x_{N-1}}v_{N-1} + \frac{x_q - x_{N-1}}{x_{N} - x_{N-1}}v_{N} & x_{N-1} < x_q
   \end{cases}    
\end{equation}
where $x \in \mathbb{R}^N$ is a strictly increasing vector of the sample points, $v \in \mathbb{R}^N$ is the values of those sample points and $x_q \in \mathbb{R}$ is the query point.
For use in the gradient expression we need the zero order hold function (with extrapolation),
\begin{equation}
\text{ZOH}(x,v,x_q) = 
\begin{cases} 
      v_1 & x_q \leq x_2 \\
      v_i & x_{i} < x_q \leq x_{i+1}, \\
      & 2 < i < N-1 \\
      v_N & x_{N} < x_q
   \end{cases}    
\end{equation}
were $x$, $v$, and $x_q$ are the same as in the linear interpolation function definition.

Consider the $k$th collocation node with input $u_k$ at time $t_k$, and control points described by time $T \in \mathbb{R}^m$ and value $U \in \mathbb{R}^m$.
Our constraint ($g$) takes the form
\begin{equation}
    g(u_k,t_k,U,T) = u_k - LI(T,U,t_k) = 0.
    \label{eq:LIConstraint}
\end{equation}
This constraint means that the the actual input $u_k$ must be equal to the interpolated input from the control points at the current time $t_k$.

The optimization method we use benefits from having analytical gradients of all constraints, which we can describe for this function.
The gradient of the constraint in \ref{eq:LIConstraint} can be found using basic calculus.
With respect to some decision variable ($y$) the gradient\footnote{In this expression the symbol $U_a^b$ is the vector of components described by $[U_a, U_{a+1}, ... ,U_{b-1}, U_{b}]$.} of this constraint is
\begin{align}
    \nabla_{y} g(u_k&,t_k,U,T) = \nabla_{y}u_k \\
     &+\text{ZOH}( T_1^{m-1}, \frac{U_2^{m} - U_1^{m-1}}{T_2^{m} - T_1^{m-1}}\nabla_{y}t_k, t_k)\\
     &+ \text{ZOH}( T_1^{m-1}, \frac{U_2^{m} - U_1^{m-1}}{T_2^{m} - T_1^{m-1}}, t_k) \text{LI}( T, \nabla_y T, t_k)\\
     &+ \text{LI}( T, \nabla_y U, t_k)
\end{align}

This constraint allows us to link all of the set point acceleration profiles together in a differentiable way.
Unfortunately the gradient is undefined at the node points themselves and is frequently discontinuous.
This discontinuous gradient can slow the optimization procedure but the constraints converge well to the desired tolerance in our testing.
A better option could be to use a piecewise cubic interpolation method to ensure the gradients are well formed, but it does not appear to be necessary. 

\subsection{Testing Simulation}
\label{sub:testingSimMethod}

To objectively test the disturbance rejection of the motions produced by the two trajectory optimization methods we implement a hybrid simulation of the ASLIP model.
The simulation uses the same model dynamics and hybrid transitions except that we treat the set point trajectory (position $r_0(t)$ and velocity $\dot{r}_0(t)$) as the inputs.
The system is forward integrated using MATLAB's variable step size ODE solver ODE45 with event sensing for the hybrid transitions.

Additionally, the leg touchdown angle must be defined for each of the motions.
The robust optimization finds an explicit time varying leg touchdown angle.
The minimum effort optimization only selects a single leg angle.
One option would be to just use that single leg angle.
A better option is to use the heuristic that the leg touchdown angle tracks a fixed horizontal touchdown location on the ground.
This policy is similar to what guinea fowl do when they encounter an unexpected step up or down \cite{Blum2014}, and should ensure that the minimum effort optimization is not unfairly hindered with an unreasonable leg angle policy.

We run this forward simulation for a set of initial condition disturbances until it reaches the next apex state.
If the disturbance is poorly handled it is possible that the body does not ever reach the next apex state.
This generally is because the model falls into the ground before lifting off or reaches liftoff with a negative vertical velocity.

\section{Results}
\label{sec:results}

To evaluate the optimization methods we generate motion plans for 625 sets of initial states and final goal states.
The experiments were run single threaded on a standard desktop computer with an Intel Core i7-7700k and 24 GB of RAM.
These correspond to every combination of five initial heights, initial horizontal velocities, final heights, and final horizontal velocities.
This results in creating motion plans for steady state gaits, changes in speed, planned step ups and planned step downs.
Observations on the optimization process and the resulting motions are presented for both the minimum effort and the explicitly robust optimizations.
Finally, the performance of the plans from both methods is tested using a separate simulation for significantly more disturbance heights than were explicitly optimized.

\subsection{Optimization Results}

\begin{figure}
  \centering
  \includegraphics[width = 0.85\columnwidth]{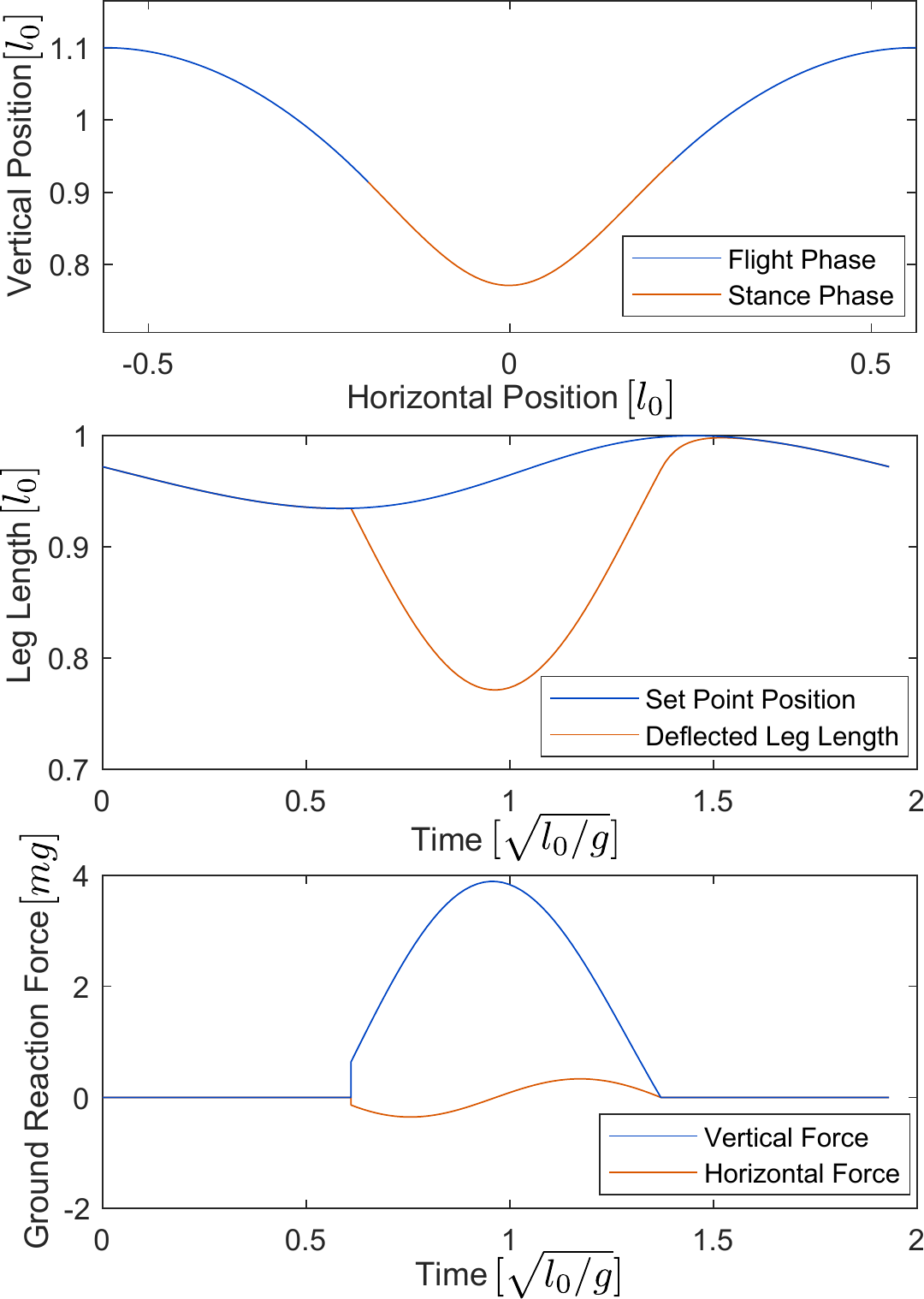}
  \caption{An example solution from the minimum effort optimization. 
  We see that the model lands with the leg retracted, then smoothly extends to the maximum length at lift off to replace the energy lost in the leg spring damping.}
\label{fig:AccelResults}
\end{figure}

The minimum effort optimizations converged to optimality relatively quickly and reliably. 
The mean solutions time was 0.90 seconds and 95\% solved in under 3.1 seconds. 
An example solution is shown in Fig. \ref{fig:AccelResults}.
The system lands with the leg set point slightly retracted and stationary.
Then throughout stance the leg extends and reaches the maximum extension just before liftoff.
As the leg extends it does positive work against the spring, replacing the energy lost in the damper during stance.
During the flight phases the leg is smoothly retracted back to prepare for the next touchdown event.
The ground reaction forces appear very similar to those seen in human running trials as well as in passive SLIP models.

In the explicitly robust optimization we used $+0.10$, $+0.05$, $-0.05$ and $-0.10$  $[l_0]$ as the errors in ground height for the disturbance cases which means this problem has five times the number of variables and constraints of the minimum effort problem.
The mean solutions time was 4.3 seconds and 95\% of successful solutions solved in under 8.9 seconds. 
This is notably slower than the minimum effort optimization, particularly when you consider the average time per iteration which was 0.01 seconds for the minimum effort and 0.2 seconds for the robust optimization.
This is to be expected because of the over five times difference in number of decision variables and constraints between the two problems.

\begin{figure}
  \centering
  \includegraphics[width = 0.85\columnwidth]{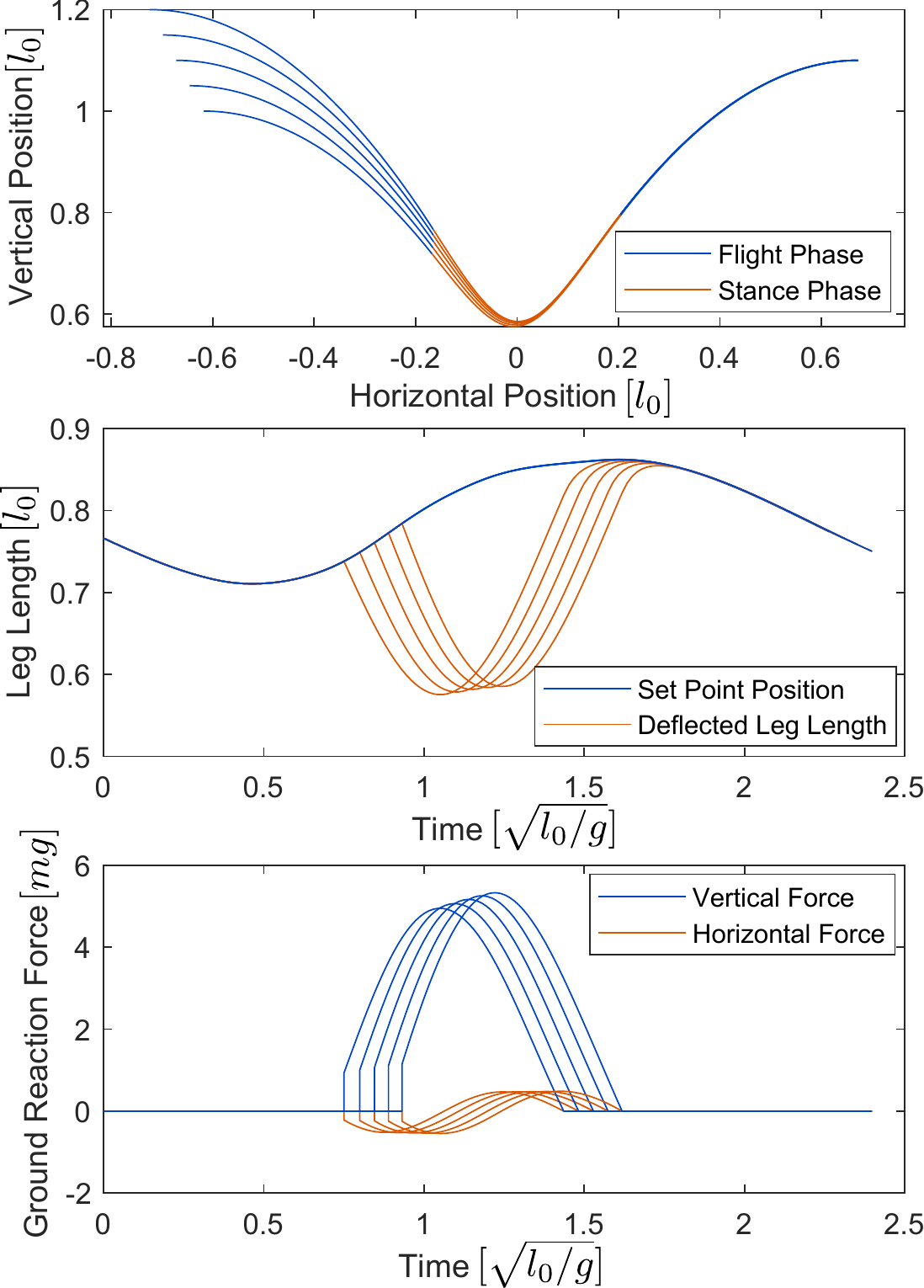}
  \caption{A robust motion plan for the actuated SLIP model. All trajectories use the same control inputs, yet they all converge to the same final apex state}
\label{fig:RobustResults}
\end{figure}

An example solution to the robust optimization is shown in Fig. \ref{fig:RobustResults}.
We can see in the top plot that all five initial heights converge back to the nearly same final height and forward velocity.
Each trajectory has a different touchdown angle, becoming steeper for later touchdowns.
The second plot shows that as the different disturbance cases touchdown, the leg is already extending in length.
All the trajectories lift off at different points in time as the leg reaches its peak extension which is well short of the maximum extension of $1$ [$l_0$].
In the ground reaction force plot at the bottom, we can see that the later the touchdown, the greater the peak force vertical force.
This intuitively makes sense because it should require a larger vertical impulse to reverse the vertical velocity of the body and return it to the final desired height. 

\subsection{Simulation Testing Results}
\begin{figure}
  \centering
  \includegraphics[width = 0.95\columnwidth]{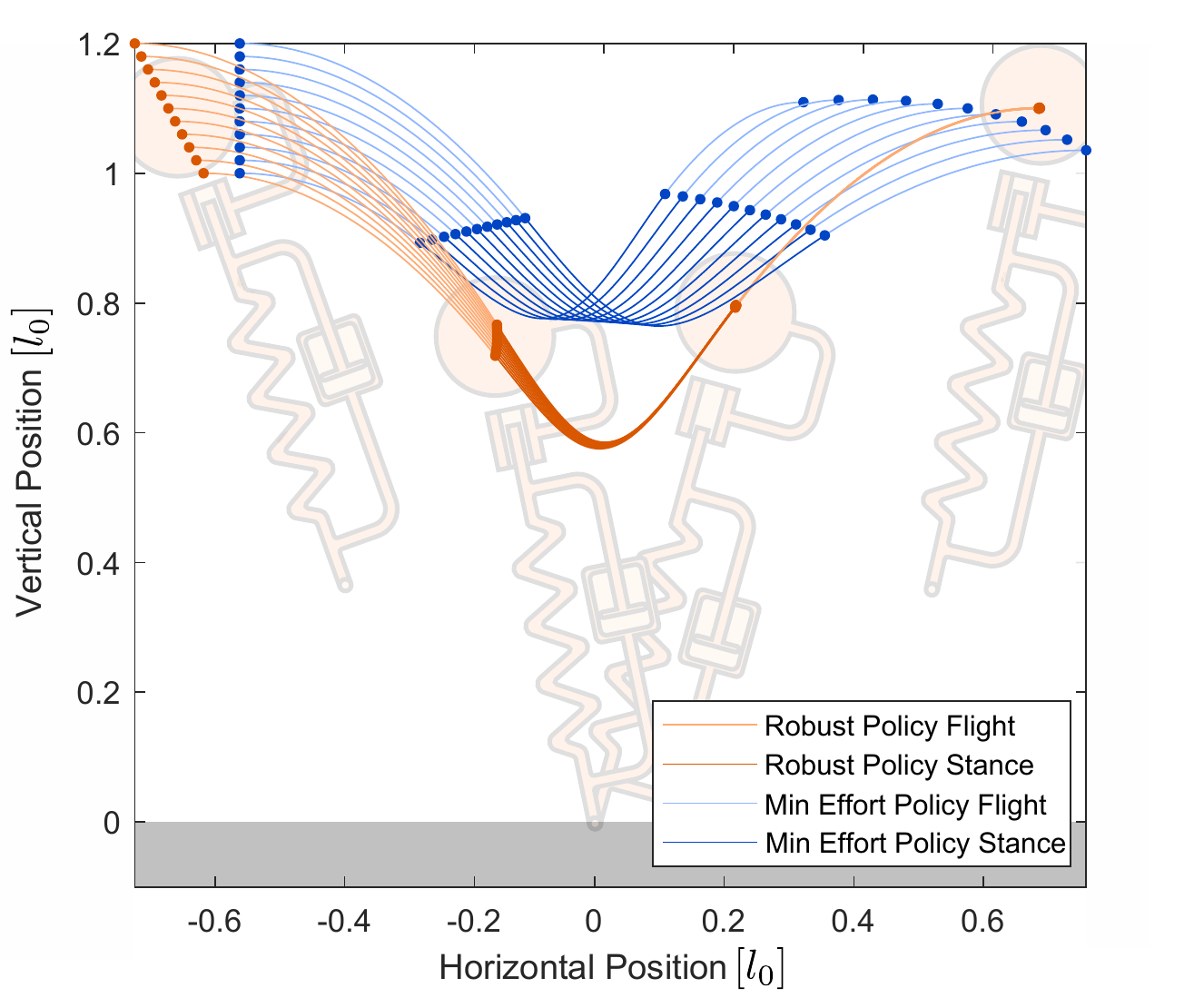}
  \caption{Simulation results of the open loop plans for the robust policy and the minimum effort policy. The robust policy (orange) has an apex height error of less than $0.001$ [$l_0$] and an apex velocity error of less than $0.001$ [$\sqrt{g l_0}$] for all disturbance cases.}
\label{fig:SimResults}
\end{figure}

The performance of the two different trajectory generators was tested using the simulation described in section \ref{sub:testingSimMethod}.
Each trajectory was tested using eleven different vertical disturbances representing different step ups and step downs, between $+0.1$ [$l_0$] and $-0.1$ [$l_0$].
An example result is shown in Fig. \ref{fig:SimResults} for a steady state gait.
Looking at the minimum acceleration results in blue, we can see that touchdown points are at similar heights because the leg set point has almost zero velocity at this point in the cycle.
As these motions exit stance, they have drastically different forward velocity but a smaller range of final heights compared to the initial disturbances.
The robust trajectories in orange show a very different reaction.
The touchdown states are in a much tighter grouping due to the leg extension and the precise leg placement policy generated by the optimization.
The states converge through stance and ascent until they reach the apex state.
All of the robust apex states in this figure have less than $0.001$ [$l_0$] error in final height and $0.001$ [$\sqrt{g l_0}$] error in final forward velocity.

Looking at the results of all of the simulations we observe that $14$\% of the disturbances caused the minimum acceleration policy to fail to reach a valid subsequent apex state because the body contacted the ground before it reached an apex state.
None of the tested disturbances caused the robust policy to fail.
When comparing the final state error of the conditions where the minimum acceleration policy did not fail, the robust policy had on average 43 times less height error and 81 times less velocity error.

\section{Conclusions}

We presented a new method to create open loop plans for an ASLIP model that are extremely robust to ground height uncertainty.
These plans are slower to calculate compared to more conventional methods, such as the minimal effort optimization presented as a standard of comparison.
The presented approach consists of optimizing many different trajectories for different disturbance cases while linking the inputs together.
The input linking used here is effective and efficient due to its analytical gradients.
The results show that the robust motions plans produced have an order of magnitude less final state error compared to the minimum effort plans.

\label{sec:conclusions}

\addtolength{\textheight}{-12cm}   




\bibliographystyle{IEEEtran.bst}
\bibliography{references.bib}

\end{document}